\documentclass[10pt,twocolumn,letterpaper]{article}

\usepackage{iccv}
\usepackage{times}
\usepackage{epsfig}
\usepackage{graphicx}
\usepackage{amsmath}
\usepackage{amssymb}

\usepackage{lipsum} 
\usepackage{graphicx}
\usepackage{grffile}
\usepackage{sidecap}
\usepackage{booktabs}
\usepackage{enumitem}
\usepackage{tabularx}
\usepackage{xfrac}
\usepackage{sectsty}
\usepackage{gensymb}
\usepackage{amsfonts,amsthm} 

\usepackage[pagebackref=true,breaklinks=true,letterpaper=true,colorlinks,bookmarks=false]{hyperref}
\iccvfinalcopy 

\def\iccvPaperID{6} 

\ificcvfinal\fi
\begin{document}

\title{Graph-Based Classification of Omnidirectional Images}

\author{Renata Khasanova \qquad Pascal Frossard\\
Ecole Polytechnique Federale de Lausanne\\
{\tt\small \{renata.khasanova, pascal.frossard\}@epfl.ch}
}

\maketitle

\begin{abstract}
	Omnidirectional cameras are widely used in such areas as robotics and virtual reality as they provide a wide field of view. Their images are often processed with classical methods, which might unfortunately lead to non-optimal solutions as these methods are designed for planar images that have different geometrical properties than omnidirectional ones. In this paper we study image classification task by taking into account the specific geometry of omnidirectional cameras with graph-based representations. In particular, we extend deep learning architectures to data on graphs; we propose a principled way of graph construction such that convolutional filters respond similarly for the same pattern on different positions of the image regardless of lens distortions. Our experiments show that the proposed method outperforms current techniques for the omnidirectional image classification problem.  
\end{abstract}

	
	
	

\section{Introduction}

Omnidirectional cameras are very attractive for various applications in robotics~\cite{bb:omni_robots,bb:omni_robots2} and computer vision~\cite{Hansen2007, Masci2014b} thanks to their wide viewing angle. Despite this advantage, working with the raw images, taken by such cameras is difficult because of severe distortion effects introduced by the camera geometry or lens optics, which has a significant impact on local image statistics. Therefore, all methods that aim at solving different computer vision tasks (e.g. detection, points matching, classification) on the images from the omnidirectional cameras need to find a way of compensating for this distortion. 

The natural way to do it is to apply calibration techniques ~\cite{Scaramuzza2006, Hughes2010, Fitzgibbon} to undistort images and then use standard computer vision algorithms. However, undistorting omnidirectional images is a non-linear operation, which requires $O(MN)$ additional operations, with $N$ and $M$ being the numbers of image pixels and images in the dataset respectively. Further, undistorting real images may suffer from interpolation artifacts. A different,  `naive', approach is to apply standard techniques directly to raw (distorted) images. However, algorithms proposed for the planar images lead to non-optimal solutions when applied to distorted images. 

One example of such standard techniques are the Convolutional Neural Networks (ConvNets)~\cite{bb:ConvNet}, which are primarily designed for regular domains~\cite{LeCun1995}. They have achieved remarkable success in various areas of computer vision~\cite{bb:Szegedy15,bb:Simonyan15,bb:Mousavian16}. 
The drawback of this solution is that ConvNets require a lot of training data for omnidirectional image classification task, as the same object will not have the same local statistics, for different image locations, which results in different filter responses. 
Therefore, the dataset should include images where same objects are seen in different parts of the image in order to reach invariance to distortions. 

\begin{figure}[t!]
	\centering
	\includegraphics[width=1\linewidth]{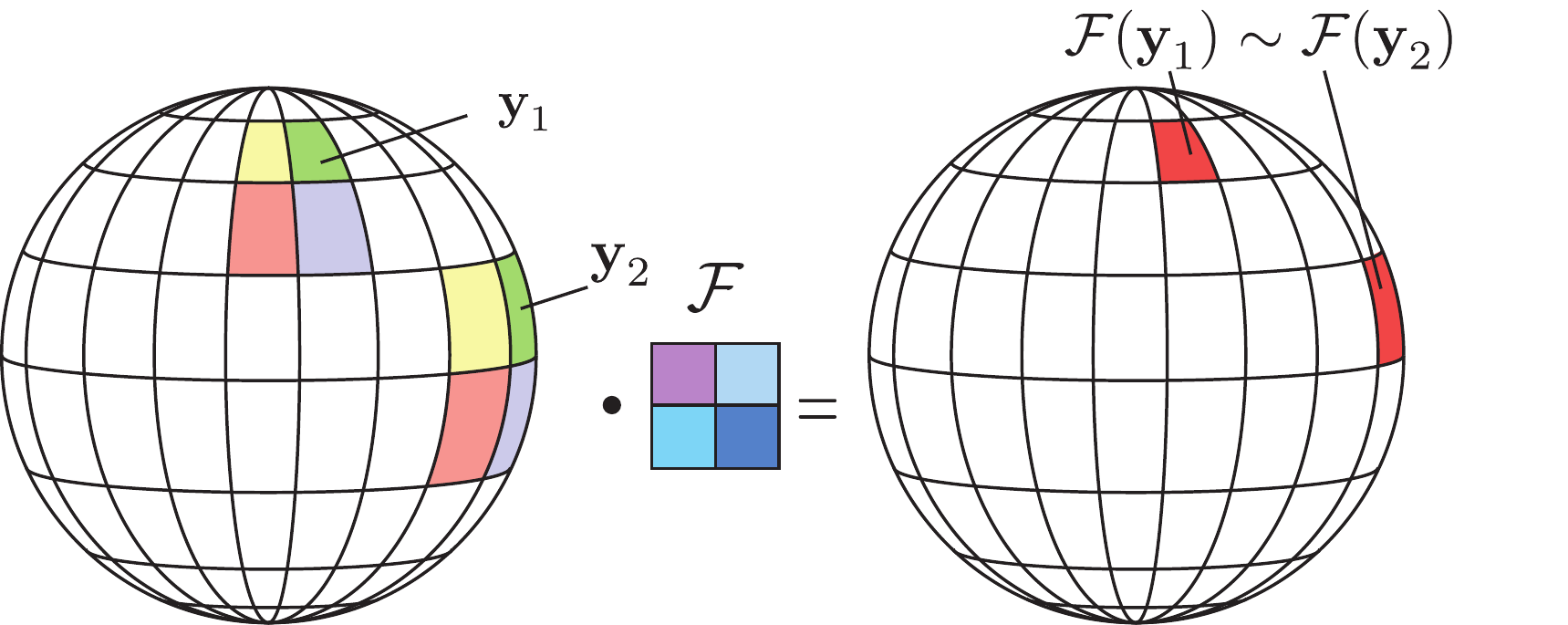}
	\caption{The proposed graph construction method makes response of the filter similar regardless of different position of the pattern on an image from an omnidirectional camera.}
	\label{fig:vis}
\end{figure}


%
%

In this work, we propose to design a solution for image classification that inherently takes into account the camera geometry. 
  Developing such a technique based on the classic ConvNets is, however, complicated due to the two main reasons. First as we mentioned before the features, extracted by the network, need to be invariant to
  positions of objects in the scene and different orientations with respect to the omnidirectional camera. Second, it is challenging to incorporate lens geometry knowledge in the structure of convolutional filters. 
  Luckily graph-based deep learning techniques have been recently introduced~\cite{Boscaini2015, bb:Mikhael, bb:TIGraNet} that allow applying deep learning techniques to irregularly structured data. 
  Our work is inspired by~\cite{bb:TIGraNet} where the authors use graphs to create isometry invariant features of images in Euclidean space. This tackles the first of the aforementioned challenges, however, the same object seen  at different positions of an omnidirectional image still remains different from the network point of view. To mitigate this issue, we propose to incorporate the knowledge about the geometry of the omnidirectional camera lens into the signal representation, namely in the structure of the graph (see Fig.~\ref{fig:vis}). 
  In summery we therefore propose the following contributions:
  \begin{itemize}
  	\setlength\itemsep{1pt}
  	\setlength{\parskip}{1pt}
  	\item a principled way of graph construction based on geometry of omnidirectional images;
  	\item graph-based deep learning architecture for the omnidirectional image classification task.
  \end{itemize}	
  

The reminder of the paper is organized as follows. We first discuss the related work in Section~\ref{s:rel_work}. Further, in Section~\ref{s:algo} we briefly introduce the TIGraNet architecture~\cite{bb:TIGraNet}, as it is tightly related to our approach, and then we describe our graph construction method that can be efficiently used by TIGraNet. Finally, we show the result of our experiments in Section~\ref{s:exp} and we conclude in Section~\ref{s:conc}.

\section{Related work}
\label{s:rel_work}

To the best of our knowledge, image classification methods designed specifically for the omnidirectional camera do not exist. Therefore, in this section we review methods designed for wide-angle cameras for different computer vision applications. Then we discuss recent classification approaches based on graphs as we believe that graph signal processing provides with powerful tools to deal with images that have an irregular structure.

%

\subsection{Wide-angle view cameras}

A broad variety  of computer vision tasks  benefit from having wide-angle cameras. For example, images from fisheye~\cite{Posada2013}, which can reach field of view (FOV) of more than $180\degree$, or omnidirectional cameras, that provide $360\degree$ FOV~\cite{Goedeme, Scaramuzza2008} are widely used in virtual reality and robotics~\cite{Goedeme, Marinho2017} applications. Despite practical their benefits these images are challenging to process due to the fact that most of the approaches are developed for planar images and suffer from distortion effects when applied to images from wide-angle view cameras~\cite{Posada2013}.

There exist different ways of acquiring an omnidirectional image. First such an image can be built based on a set of multiple images, taken either by a camera that is rotated around its center of projection, or by multiple calibrated cameras. Rotating camera systems, however cannot be applied to dynamic scenes, while multi-camera systems suffer from calibration difficulties. Alternatively, one can obtain an omnidirectional image from dioptric or catadioptric cameras~\cite{Tosi}. 
Most of the existing catadioptric cameras have the following mirror types: eliptic, parabolic and hyperbolic. The authors in~\cite{Geyer2001} show that such mirror types allow for a bijective mapping of the lens surface to a sphere, which simplifies processing of omnidirectional images. 
In our paper we work with this spherical representation of catadioptric omnidirectional cameras.
The analysis of images from wide-angle cameras remains however an open problem.

For example, the standard approaches for interest point matching propose affine-invariant descriptors such as SIFT~\cite{Lowe1999}, GIST~\cite{Torralba2003}. However, designing descriptors that preserve invariance to geometric distortions for wide-angle camera's images is challenging.
One of the attempts to achieve such invariance is proposed by~\cite{Rituerto2014}, where the authors extend the GIST descriptor to omnidirectional images by exploiting their circular nature. Instead of using hand-crafted descriptors, the authors in~\cite{Strecha2012} suggest to learn them from the data by creating a similarity preserving hashing function. Further, inspired by the aforementioned method, the work in~\cite{Masci2014b} proposes to learn descriptors for images from omnidirectional cameras using a siamese neural network~\cite{Hadsell}. While this method is not using specific geometry of the lens, it significantly outperforms state-of-the-art as it encodes transformations that are present in the omnidirectional images. However, the method requires carefully constructed training dataset to learn all possible variations of the data.

Contrary to the previous approaches, the methods in~\cite{Hansen2007, Cruz-Mota2012, Hadj-Abdelkader2008} design a scale invariant SIFT descriptor for the wide-angle cameras based on the result of the work in \cite{Geyer2001} that introduced a bijection mapping between omnidirectional images and a spherical surface. In particular, the method in~\cite{Hansen2007}  maps images to a unit sphere, and those in~\cite{Cruz-Mota2012} propose two SIFT-based algorithms, which work in spherical coordinates. The first approach (local spherical) matches points between two omnidirectional images, while the second one (local planar) works between spherical and planar images. 
Finally, the authors in~\cite{Hadj-Abdelkader2008} adapt a Harris interest point detector~\cite{Harris1988} to spherical nature of images from omnidirectional cameras. All the aforementioned works are designed for interest point matching task. In our work we use the similar idea of mapping omnidirectional images to the spherical surface~\cite{Geyer2001} for omnidirectional image classification problem.

Omnidirectional cameras have also been widely uses in other computer vision and robotics tasks. For example, the authors in~\cite{Bogdanova2007} propose a segmentation method for catadioptric cameras. They derive explicit expression for edge detection and smoothing differential operators and design a new energy functional to solve segmentation problem. The work in \cite{Arican} then develops a stereo-based depth estimation approach from multiple cameras. Further, the authors in \cite{Tosi} extend previous geometry-based calibration approach to compute depth and disparity maps from images captured by a pair of omnidirectional cameras. They also suggest an efficient way of sparse 3D scene representation. The works in~\cite{Goedeme, Scaramuzza2008} then use omnidirection cameras for robot self-localisation and reliable estimation of the 3D map of the environment~(SLAM). The authors in~\cite{Tosic} propose a motion estimation method for catadioptric omnidirectional images by the evaluation of the correlation between them from arbitrary viewpoints. Finally, the work in~\cite{DeSimone} utilizes geometry of the omnidirectional camera to adapt the quantization tables of ordinary block-based transform codecs for panoramic images computed by equirectangular projection.

In summary, processing images from omnidirectional cameras becomes an important topic in the computer vision community. However, most of the existing solutions rely on methods developed for planar images. In this paper we are particularly interested in image classification tasks and propose a solution based on a combination of powerful deep learning architecture and camera lens geometry.

\subsection{Deep learning on graphs}

We briefly review here classification methods based on deep learning algorithms (DLA) for graph data as DLA has proven its efficiency in many computer vision tasks and graphs allow extending these methods to irregularly structured data, such as omnidirectional images. A more complete review can be found in  \cite{bb:BronsteinBLSV16}.

First, the authors in~\cite{Bruna2013, Henaff2015} propose a new deep learning architecture that uses filters in spectral domain to work with irregular data, where they add a smoothing constraint to avoid overfitting. 
These methods have high computational complexity as they require eigendecomposition as a preprocessing step.  
To reduce this complexity,  the authors in \cite{bb:Mikhael} propose using Chebyshev polynomials, which can be efficiently computed in an iterative manner and allow for fast filter generation. The work in~\cite{Kipf2016}  uses similar polynomial filters of degree 1, which allow training deeper and more efficient models for semi-supervised learning tasks without increasing the complexity.

Finally, the recent method in~\cite{bb:TIGraNet} introduces graph-based global isometry invariant features. Their approach is developed for regular planar images. 
We, on the other hand, propose to design a graph signal representation that decreases feature sensitivity to different types of geometric distortion introduced by omnidirectional cameras.

\newcommand{\bT}{\textbf{T}}
\newcommand{\bp}{\textbf{x}}
\newcommand{\bX}{\textbf{X}}
\newcommand{\bS}{\textbf{S}}
\newcommand{\by}{\textbf{y}}

\section{Graph convolutional network}
\label{s:algo}

In this paper we construct a system to classify images from omnidirectional camera based on a deep learning architecture. In particular, we extend the network from \cite{bb:TIGraNet} to process images with geometric distortion. In this section we briefly review the main components of this approach. The system in \cite{bb:TIGraNet} takes as input images that are represented as signals on a grid graph and gives classification labels as output. Briefly this approach proposes a network of alternatively stacked spectral convolutional and dynamic pooling layers, which creates features that are equivariant to the isometric transformation. Further, the output of the last layer is processed by a statistical layer, which makes the equivariant representation of data invariant to isometric transformations. Finally, the resulting feature vector is fed to a number of fully-connected layers and a softmax layer, which outputs the probability distribution that the signal belongs to each of the given classes. 

We extend this transformation-invariant classification algorithm to omnidirectional images by incorporating the knowledge about the camera lens geometry in the graph structure. We assume that the image projection model is known and propose representing 
images as signals $\by$ on the irregular grid graph $G$. More formally, the graph is a set of nodes, edges and weights. Thus, each graph signal $\by: \{\by(v_i)\}$ is defined on nodes $v_i, i \in [1..N]$ of~$G$.  We denote by $\textbf{A}$ an adjacency matrix of $G$, which shows weighted connection between vertices, and by $\textbf{D}$ a diagonal degree matrix with $D_{ii} = \sum_{j=1}^N A_{ij}$. This allows us to  define Laplacian matrix\footnote{We use non-normalized version of Laplacian matrix which is different from \cite{bb:TIGraNet} to simplify the derivation. The same result can be obtained with the normalized version of $\textbf{L}$.} as follows:
\begin{equation}
\label{eq:lapl}
 \textbf{L} = \textbf{D} - \textbf{A}.
\end{equation}
The Laplacian matrix is an operator, that is  widely used in graph signal processing~\cite{shuman2013emerging}, because it allows to define a graph Fourier transform to perform analysis of graph signals. The transformed signal reads:
\begin{equation}
\label{eq:fourier}
\hat{\by}(\lambda_l) :=\langle \by, \textbf{u}_l\rangle,
\end{equation}
where $\lambda_l$ is an eigenvalue of $\textbf{L}$ and $\textbf{u}_l$ is the associated eigenvector. It gives a  spectral representation of the signal $\by$ and $\lambda_l, l \in [0..N-1]$ provide a similar notion of frequencies as in classical Fourier analysis. 
Thus, the filtering of a graph signal $\by$ can be defined on graphs in the spectral domain:
\begin{equation}
\label{eq:filter_spectral}
\mathcal{F} (\by(v_i))  = \sum_{l=0}^{N-1} \hat{f}(\lambda_l) \hat{\by}(\lambda_l) \textbf{u}_l(v_i),
\end{equation}
where  $\mathcal{F} (\by(v_i))$ is the filtered signal value on the node $v_i$ and $\hat{f}(\lambda_l)$ is a graph filter. Graph filters can be constructed as polynomial function, namely $\hat{f}(\lambda_l) = \sum_{m=0}^M \alpha_m \lambda_l^m$, where $M$ is the degree of the polynomial and $\alpha_{m}$ are the parameters. Such filters can advantageously be applied directly in vertex domain to avoid computationally expensive eigendecomposition~\cite{shuman2013emerging}:
\begin{equation}
\label{eq:filter_resp}
\mathcal{F} (\by) =  \left[\sum_{m=0}^M \alpha_{m} \textbf{L}^m\right]^T \by,
\end{equation}
where the filter has the form $\mathcal{F} = \sum_{m=0}^M \alpha_{m} \textbf{L}^m$.
 The spectral convolutional layer in the deep learning system of~\cite{bb:TIGraNet} consists of $J$ such filters $\mathcal{F}_j, j\in [1..J]$.
Each column $i$ of $\mathcal{F}_j$ represents localization of the filter on the node $v_i$. These nodes are chosen by preceding pooling layer, which selects nodes with maximum response. 
Finally, a statistical layer collects global multi-scale statistic of input feature maps, which results in rotation and translation invariant features. Please refer to~\cite{bb:TIGraNet} for more details about the TIGraNet architecture.

 

In the next section, we show how this architecture can be extended for omnidirectional images. In particular, we discuss how to compute the weights $\textbf{A} : \{w_{ij}\}$ between the nodes according to the lens geometry in order to build a proper Laplacian matrix for spectral convolutional filters.

\section{Graph-based representation}

\subsection{Image model}
\label{s:im_mod}

\begin{figure}[t!]
	\centering
	\includegraphics[width=0.7\linewidth]{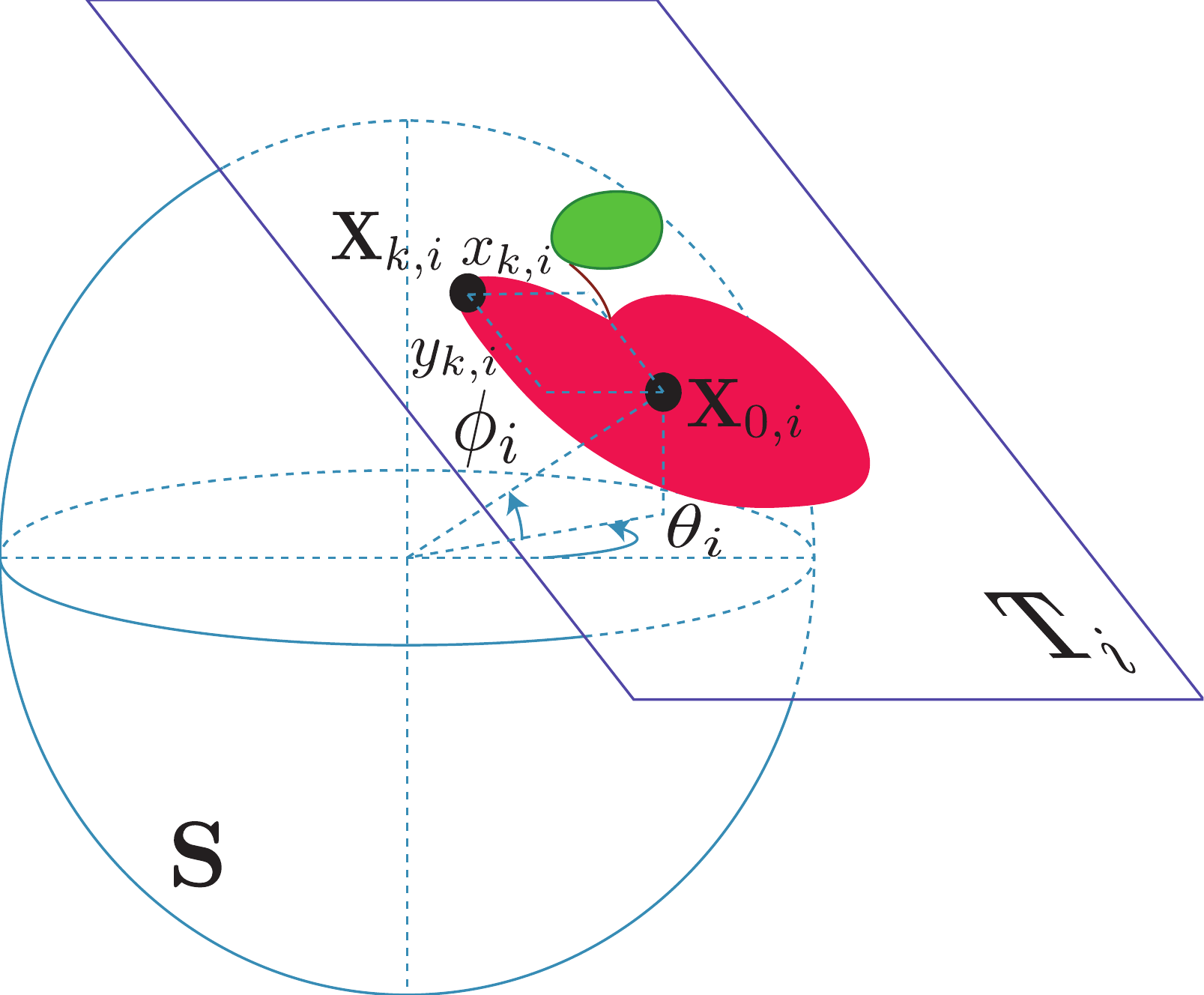}
	\caption{Example of the gnomonic projection. An object from tangent plane $\bT_i$ is projected to the sphere at tangency point $\bX_{0,i}$, which is defined by spherical coordinates $\phi_i, \theta_i$. The point $\bX_{k,i}$ is defined by coordinates $(x_{k,i}, y_{k,i})$ on the plane. }
	\label{fig:gnom_proj}
\end{figure}

An omnidirectional image is typically represented as a spherical one (see Fig.~\ref{fig:gnom_proj}), where each point $\textbf{X}_k$ from 3D space is projected to the points $\textbf{x}_k$ on the spherical surface $\bS$ with radius $r$, which we set to $r=1$ without loss of generality. The point $\textbf{x}_k$ is then uniquely defined by its longitude $\theta_k \in [-\pi, \pi]$ and latitude $\phi_k \in [-\frac{\pi}{2}, \frac{\pi}{2}]$ and its coordinates can be written as:
\begin{equation}
 \\
 \textbf{x}_k :\\
 \\
\begin{bmatrix}
\cos\theta_k \cos\phi_k \\
\sin\theta_k \cos\phi_k \\
\sin\phi_k \\
\end{bmatrix}, k \in [1..N].
\end{equation}
\noindent
We consider objects on a plane that is tangent to the sphere $\bS$. We denote by $\textbf{X}_{k,i}$ a 3D space point on the plane $\textbf{T}_i$ tangent to the sphere at $(\phi_i, \theta_i)$. The point $\textbf{X}_{k,i}$ is defined by the coordinates $(x_{k,i}, y_{k,i})$ on the plane $\textbf{T}_i$.
We, further, denote by $\bp_{k,i}: (\phi_k, \theta_k)$ the points on the surface of the sphere that are obtained by connecting its center and the point $\textbf{X}_{k,i}$ on $\textbf{T}_i$. 
We can find coordinates of each point $\textbf{X}_{k,i}$ on $\bT_i$ by using the gnomonic projection, which generally provides a good model for omnidirectional images \cite{bb:map_projection}:
\begin{equation}
\label{eq:gnom}
\begin{array}{l}
x_{k,i}=\frac{\cos\phi_k \sin(\theta_k-\theta_i)}{\cos c}, \\
\\
y_{k,i}=\frac{\cos\phi_i \sin \phi_k -\sin \phi_i \cos\phi_k \cos(\theta_k - \theta_i)}{\cos c}, \\
\end{array}
\end{equation}
where $c$ is the angular distance between the point $(x_{k,i},y_{k,i})$ and the center of projection $\bX_{0,i}$ and is defined as follows:
\begin{equation}
\begin{array}{l}
\cos c = \sin\phi_i \sin \phi_k + \cos \phi_i \cos\phi_k \cos(\theta_k - \theta_i), \\
c = \tan^{-1} \left(\sqrt{x_{k,i}^2 + y_{k,i}^2} \right).  
\end{array}
\end{equation}
%

\noindent
Fig.~\ref{fig:gnom_proj} illustrates an example of this gnomonic projection. 

\begin{figure}[t!]
	\centering
	\includegraphics[width=0.7\linewidth]{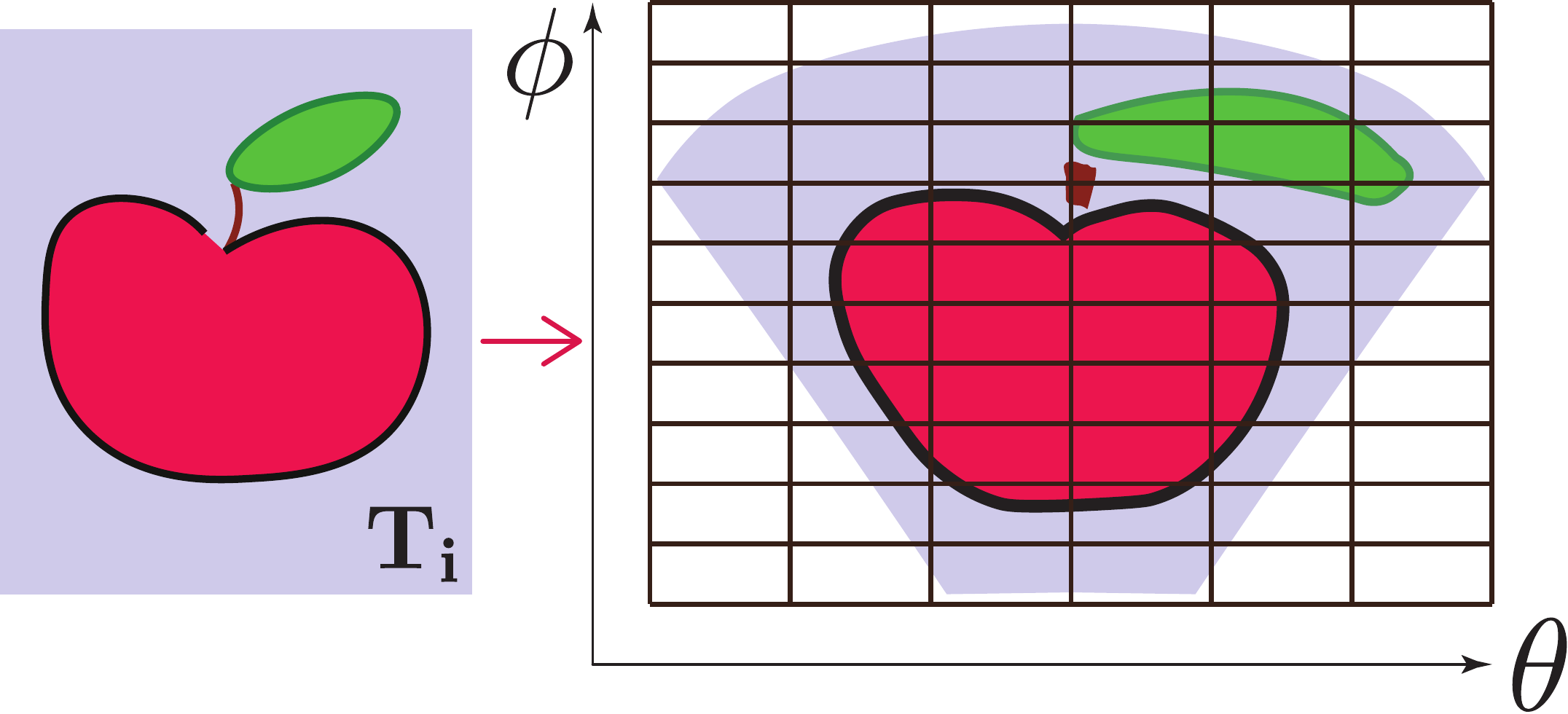}
	\caption{Example of the equirectangular representation of the image. On the left, the figure depicts the original image on the tangent plane $\bT_i$, on the right, projected to the points of the sphere. To build an equirectangular image the values points on the discrete regular grid are often approximated from the values of projected points by interpolation.}
	\label{fig:equirect}
\end{figure}

In order to easily process the signal defined on the spherical surface, it is typically projected to an equirectangular image (see Fig.~\ref{fig:equirect}). The latter represents the signal on the regular grid with step sizes $\Delta \theta $ and $ \Delta \phi$ for angles $\theta$ and $\phi$ respectively. In this paper we work with these equirectangular images and assume that the object, which we are classifying, is lying on a plane $\bT_i$ tangent to the sphere $\bS$ at the point $(\phi_i, \theta_i)$. Our work could however be adapted to other projection models, such as \cite{Miyamoto1964}. Finally, each point on the equirectangular image is considered as a vertex $v_k$ in our graph representation. The graph then connects nearest neighbors of the equirectangular image $\by(v_i)=\by(\phi_i, \theta_i)$ 

\vspace{0.1cm}

\subsection{Weight design}

Our goal is to develop a transformation invariant system, which can recognize the same object on different planes $\bT_i$ that are tangent to $\bS$ at different points $(\phi_i,\theta_i)$ without any extra training. The challenge of building such a system is to design a proper graph signal representation that allow compensating for the distortion effects that appear on different elevations of $\bS$. In order to properly define the structure, namely to compute the weights that satisfy the above condition we analyze, how a pattern projected a plane $\bT_e$ at equator $(\phi_e=0, \theta_e=0)$ varies on $\bS$ with respect to the same pattern projected onto another plane $\bT_i$ tangent to the sphere at $(\phi_i, \theta_i)$. We use this result to minimize the difference between filter responses of two projected pattern versions. Generally, the weight choice depends on distances $d_{ij}$ between neighboring nodes of graph $w_{ij} = g(d_{ij})$. In this section we show that the function $g(d_{ij}) = \frac{1}{d_{ij}}$ satisfies the above invariance condition.



\begin{figure}
	\centering
	\begin{tabular}{ccc} 	
		\includegraphics[width=0.25\linewidth]{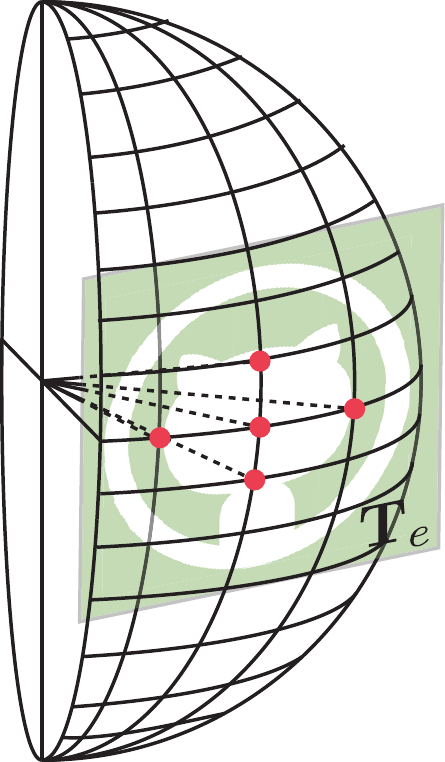} &
		\includegraphics[width=0.25\linewidth]{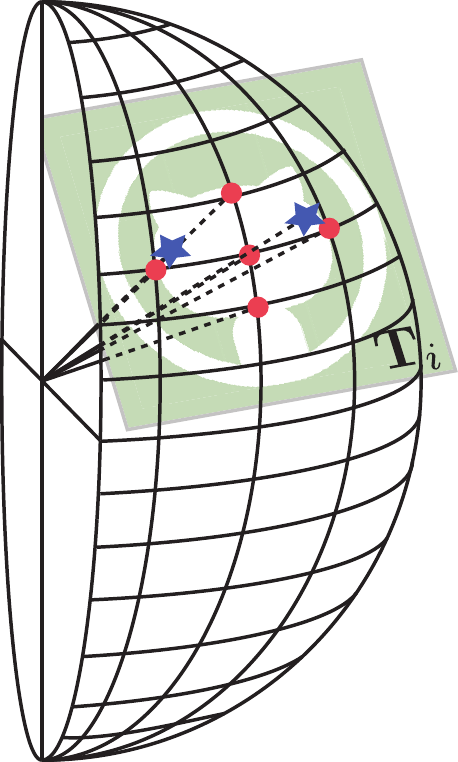} &
			\includegraphics[width=0.4\linewidth]{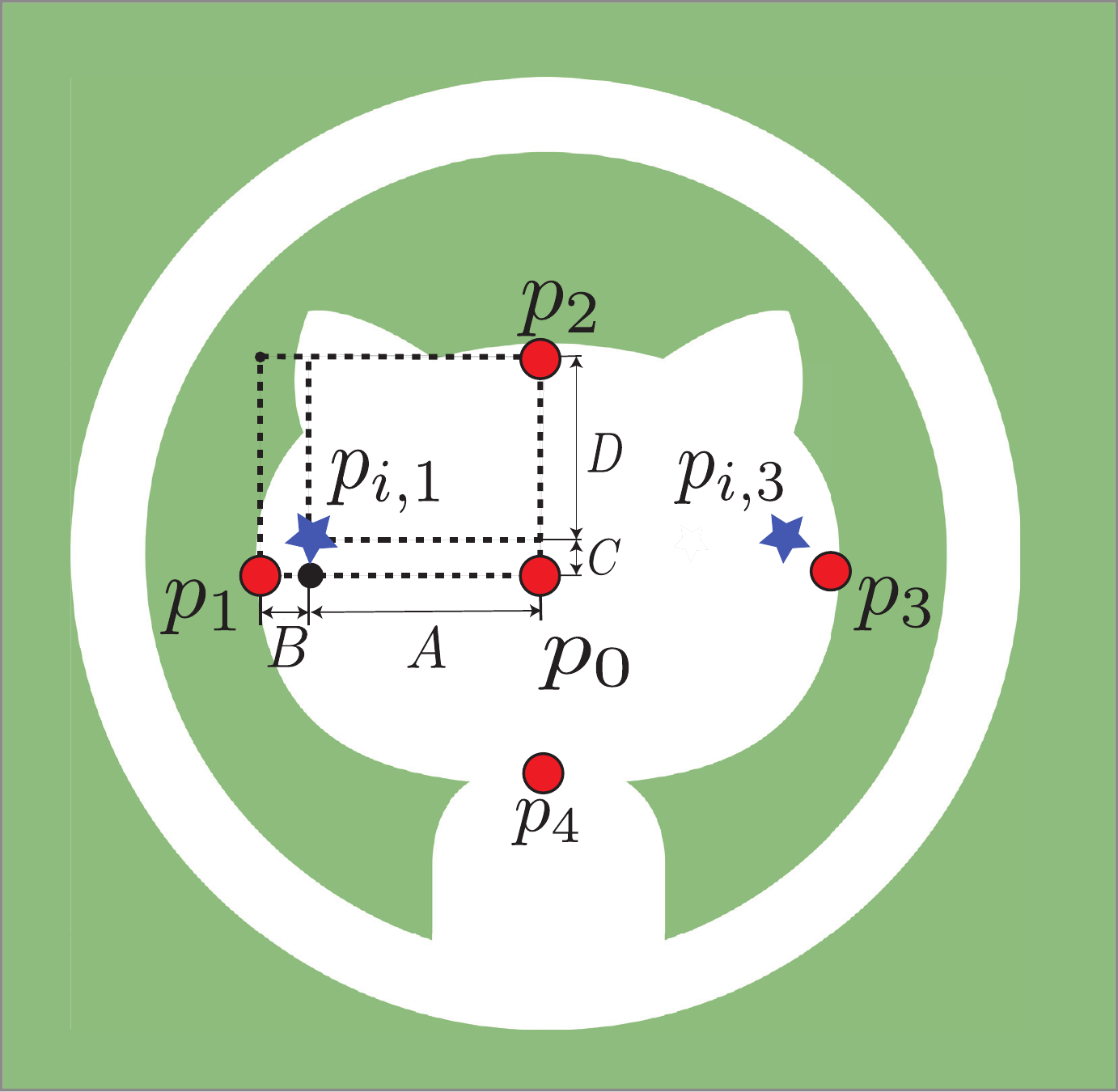} \\
			a) & b) & c) \\
	\end{tabular}
	\caption{a) We choose pattern $p_{0}, .., p_{4}$ from an object on tangent plane $\bT_e$ at equator $(\phi_e=0, \theta_e=0)$ (red points) and then, b) move this object on the sphere by moving the tangent plane $\bT_i$ to  point $(\phi_i, \theta_i)$. c) Thus, the filter localized at tangency point $(\phi_i, \theta_i)$ uses values $p_{i,1}, p_{i,3}$ (blue points) which we can obtain by interpolation.}
	\label{fig:pattern}
\end{figure}

\paragraph{Pattern choice.}
For simplicity we consider a 5-point pattern $\{p_0, \dots, p_4\}$ on a tangent plane, which is depicted by the Fig.~\ref{fig:pattern}:
\begin{equation}
\label{eq:equator_equalities}
p_j := \bX_{j,e}, \quad \forall j \in [0..4],
\end{equation}
where $\bX_{j,e}$ are the points on the plane $\bT_e$ tangent to an equator point $\phi_e=0, \theta_e=0$ and $\bX_{0,e} = \bp_{0,e}$ is the tangency point.
Further, pattern points $\{p_0, \dots, p_4\}$ are also chosen in such a way that they are projected to the following locations on the sphere $\bS$: \begin{equation}
\label{eq:pattern}
\begin{array}{ll}
p_0 &\mapsto (0, 0) \\
p_2, p_4 &\mapsto (0 \pm \Delta \phi, 0)  \\
p_1, p_3 &\mapsto (0, 0 \pm \Delta \theta)  \\
\end{array}.
\end{equation}
\noindent
These essentially correspond to the pixel locations of the equirectangular representation of the spherical surface introduced in Section~\ref{s:im_mod}. The chosen pattern has the following coordinates on the tangent plane at equator $\bT_e$ :
\begin{equation}
\label{eq:sphere_point_equator}
\begin{array}{llcl}
\bX_{0,e} &  =(0,0) \\
\bX_{2,e},\bX_{4,e} & = (0,\pm \tan \Delta \phi) \\
\bX_{1,e},\bX_{3,e} &  = \left(\pm \tan \Delta \theta, 0 \right) \\
\end{array}.
\end{equation}

\paragraph{Filter response.}
Our objective is to design a graph, which can encode the geometry of an omnidirectional camera in the final feature representation of an image. Ideally, the same object at different positions on the sphere should have the same feature response (see Fig.~\ref{fig:vis}) or equivalently they should generate the same response to given filters. Therefore, we choose the graph construction, or equivalently the weights of the graph in such a way that the difference between the responses of a filter applied to gnomonic projection of the same pattern on different tangent planes $\bT_i$ is minimized. We consider a graph
 where each node is connected with $4$ of its nearest neighbours and take as an example the polynomial spectral filter $\mathcal{F} = \textbf{L}$ of degree $1$ ($\alpha_{0,j}=0, \alpha_{1,j}=1$), we can compute the filter response according to the Eq.~(\ref{eq:lapl}) and Eq.~(\ref{eq:filter_resp}):
\begin{equation}
\mathcal{F}(\by(v_i)) = D_{ii} \by(v_i) - \sum_{j\in \mathcal{E}} A_{ij} \by(v_j),
\end{equation}
at the vertex $p_0$, one can write in particular:
\begin{equation}
\label{eq:filter_response_equator_}
\begin{array}{rl}
\mathcal{F}(\by(p_0)) =& 2(w_V + w_H) \by(p_0) - w_V (\by(p_2) + \by(p_4)) \\
&- w_H (\by(p_1) + \by(p_3)),
\end{array}
\end{equation}
where $w_V, w_H$ are the weight of the `vertical' and `horizontal' edges of the graph. For the graph nodes representing points $p_l: (\theta_l, \phi_l)$ and $p_m: (\theta_m, \phi_m)$ we refer to edges as `vertical' or `horizontal' if $\theta_l=\theta_m, \phi_l \neq \phi_m$ or $\theta_l\neq\theta_m, \phi_l = \phi_m$ correspondingly.

We now calculate the filter response $\mathcal{F}(\by(p_{0,i}))$ for a point $p_0$ on the tangent plane $\bT_i$ and compare the result with Eq.~(\ref{eq:filter_response_equator_}).
For simplicity, we assume that we shift the position of the tangent plane by an integer number of pixel positions on the spherical surface, namely $\phi_i, \theta_i$ corresponds to a node of the graph given by the equirectangular image.

According to the gnomonic projection (Eq.~(\ref{eq:gnom})), the locations of $p_{k,i}, k=[0, .., 4]$ defined on the surface of $\bS$ as $(\phi_i, \theta_i), (\phi_i \pm \Delta \phi, \theta_i), (\phi_i, \theta_i \pm \Delta \theta)$ correspond to the following positions $\bp_{k,i}=(x_{k,i}, y_{k,i})$ on the tangent plane $\bT_i$:
\begin{equation}
\label{eq:sphere_point_general}
\begin{array}{ll}
\bX_{0,i} &  = (0,0) \\
\bX_{2,i},\bX_{4,i} &  = (0,\pm \tan \Delta \phi)\\
\bX_{1,i},\bX_{3,i} & = \left(\pm \frac{\cos \phi_i \sin \Delta \theta}{\sin^2\phi_i +\cos ^2 \phi_i \cos \Delta \theta},  \frac{\sin\phi_i\cos\phi_i(1-\cos\Delta\theta)}{\sin^2\phi_i +\cos ^2 \phi_i \cos \Delta \theta}\right) \\
\end{array}
\end{equation}

The tangent plane's positions of points $\bX_{0,i},\bX_{2,i},\bX_{4,i}$ are independent of $(\phi_i, \theta_i)$, therefore their values remain the same as those of $p_0$, $p_2$ and $p_4$ respectively. However, the positions of points $\bX_{1,i},\bX_{3,i}$ depend on $(\phi_i,\theta_i)$, so that we need to interpolate the values of the pattern signal at the vertices $p_{i,1}$ and $p_{i,3}$ (see Fig.~\ref{fig:pattern}). 

We can approximate the values at $p_{i,1}$ and $p_{i,3}$ using the bilinear interpolation method~\cite{bb:bi_li_int}. We denote by $A,B,C,D$ the distances between the corresponding points of the pattern $\bT_i$, as shown in Fig.~\ref{fig:pattern}~(c). We can then express $p_{i,1}$ and $p_{i,3}$ as:
\begin{equation}
	\label{eq:v_prime}
	\begin{array}{c}
		\by(p_{i,1})
		 = E^{-1}(AD \by(p_1)+ BD \by(p_0)+CB \by(p_2)), \\
		\by(p_{i,3}) = E^{-1} (AD \by(p_3)+ BD \by(p_0)+CB \by(p_2)), \\
	\end{array}
\end{equation}
where, using Eq~(\ref{eq:sphere_point_general}),
\begin{equation}
\begin{array}{rcl}
E &=& (C+D)(A+B), \\
A+B &=& \tan \Delta \theta, \\
C+D &=& \tan \Delta \phi \\
\end{array}.
\end{equation}
\noindent
Using Eq.~(\ref{eq:v_prime}) we can then write the expression for the filter response $\mathcal{F}(\by(p_{0,i}))$, as follows:
\begin{equation}
\label{eq:filter_response_prime}
\begin{array}{rl}
\mathcal{F} (\by(p_{0,i})) &= 2(w_{i,V}+ w_{i,H}) \by(p_{0}) \\&- w_{i,V}(\by(p_{2}) + \by(p_{4})) \\&- w_{i,H} (\by(p_{i,1})+ \by(p_{i,3})),
\end{array}
\end{equation}
where $w_{i,H}$ and $w_{i,V}$ are the weights of the `horizontal' and `vertical' edges of the graph at points with the elevation $\phi_i$.

\paragraph{Objective function.}
We now want the filter responses a $p_0$ and $p_{0,i}$ in~(Eq.~(\ref{eq:filter_response_equator_}) and Eq.~(\ref{eq:filter_response_prime})) to be close to each other in order to build translation-invariant features. Therefore,  we need to find weights $w_H, w_V, w_{i,H}$ and $w_{i,V}$ such that the following distance is minimized:
\begin{equation}
\label{eq:difference}
\left|\mathcal{F} (\by(p_{0,e}))  - \mathcal{F} (\by(p_{0,i})) \right|.
\end{equation}

\noindent
Additionally, as we want to 
build a unique graph independently of the tangency point of $\bT_i$ and $\bS$, we have additional constraint of $w_V=w_{i,V}$. The latter is important, as from Eq.~(\ref{eq:sphere_point_general}) we can see that `vertical' (or elevation) distances are not affected by translation of the tangent plane.

We assume that the camera has a good resolution, which leads to $\Delta \theta \simeq 0$. Therefore, based on Eq.~(\ref{eq:difference}), we can derive the following:
\begin{equation}
\begin{cases}
w_H \simeq \left(\frac{\cos \phi_i \cos \Delta \theta}{\sin^2\phi_i + \cos^2\phi_i\cos\Delta\theta}\right) w_{i,H}  = w_{i,H} \cos \phi_i,\\
\cos \Delta \theta \simeq 1. \\
\end{cases}
\end{equation}

Therefore, under our assumptions, we can conclude that the difference between filter responses, defined by Eq.~(\ref{eq:difference}) is minimized if the following condition is valid:
\begin{equation}
\label{eq:weight_rel}
w_{i,H} = w_{H}\left(\cos\phi_i\right)^{-1},
\end{equation}
\noindent
where $w_H$ is the weight of the edge between points on the equator of the sphere $\bS$. 

Now, we can use this result to choose a proper function $g(d_{ij})$ to define the weights $w_{i,H}$ based on the Euclidean distances between two neighboring points $(\bp_i,\bp_j)$ on the sphere $\bS$.
For the case when $\phi_i=\phi_j=\phi_*, \theta_i\neq\theta_j$ the Euclidean distance can be expressed as follows:
\begin{equation}
\label{eq:distance_formula}
d^2_{ij} = r^2 (1- \cos\Delta\theta)(1+\cos2\phi_*)= r^2 \cos^2\phi_*.
\end{equation}
\noindent
For simplicity let us denote $d_{ij} = d_{\phi_*}$, where $\phi_*$ is the elevation of the points $\bp_i, \bp_j$. Using these notations, we can compute the proportion between distances $d_{\phi_e}$ and $d_{\phi_*}$, which are the distances between neighboring points at equator $\phi_e=0$ and elevations $\phi_*$ respectively. It reads:
\begin{equation}
\label{eq:distance_rel}
\frac{d_{\phi_*}}{d_{\phi_{e}}} = \frac{\cos\phi_*}{\cos\phi_{e}}.
\end{equation}
Given Eq.~(\ref{eq:weight_rel}), we can rewrite Eq.~(\ref{eq:distance_rel}) for elevation $\phi_*=\phi_i$ as:
\begin{equation}
\cos\phi_i = \frac{d_{\phi_i}}{d_{\phi_e}} = \frac{w_H}{w_{i,H}}.
\end{equation}
\noindent
As we can see, the distance between neighboring points on different elevation levels $\phi_i$, is proportional to $\cos \phi_i$. 
Given Eq.~(\ref{eq:difference}), we can see that making weights inversely proportional to Euclidean distance allows to minimize difference between filter responses. Therefore, we propose using $w_{i,H}$ as:
\begin{equation}
w_{i,H} = \frac{1}{d_{\phi_i}}.
\end{equation}

This formula can also be used to compute the weights for vertical edges, as the distance $d$ between any pair neighboring points $(\bp_i,\bp_j)$, for which $\theta_i = \theta_j$ and $\phi_i \neq \phi_j$ is constant. This nicely fits with our assumption that the weights of `vertical' edges should not depend on the tangency point of plane $\bT_i$  and sphere $\bS$. 

Thus, summing it up we choose the weights $w_{ij}$ of a graph based on the Euclidean distance between  pixels on spherical surface $d_{ij}$ as follows:
\begin{equation}
w_{ij} = \frac{1}{d_{ij}}.
\end{equation}
The graph representation finally forms the set of signals $\by$ that are fed into the network architecture defined in Section~\ref{s:algo}.



\section{Experiments}
\label{s:exp}

In this section we present our experiments. We first describe the datasets that we use for evaluation of our algorithm. We then compare our method to state-of-the-art algorithms.

We have used the following two datasets for the evaluation of our approach.

	\textbf{MNIST-012} is based on a fraction of the popular MNIST dataset~\cite{bb:MNIST} that consists of $1100$ images of size $28 \times 28$, subdivided in three different digit classes: `0', `1' and `2'. We then randomly split these images into training, validation and test sets of $600$, $200$ and $300$ images respectively. In order to make this data suitable for our task we project them to the sphere at a point $(\phi_i, \theta_i)$, as depicted by Fig.~\ref{fig:gnom_proj}. To evaluate accuracy with the change of $(\phi_i, \theta_i)$, for each image we randomly 
	sample from 9 different positions: $\phi_i \in \{0, 1/8, 1/4\}, \theta_i \in \{\pm 1/8, 0\}$. Finally we compute equirectangular images (see Fig.~\ref{fig:gnom_proj}) from these projections, as defined in Section~\ref{s:im_mod} and use the resulting images to analyze the performance of our method.

	\textbf{ETH-80} is a modified version of the dataset introduced in~\cite{bb:ETH80}. It comprises $3280$ images of size $128 \times 128$ that features $80$ different objects from $8$ classes, each seen from $41$ different viewpoints. We further resize them to $50 \times 50$ and randomly split these images into $2300$ and $650$ training and test images, respectively. We use the remaining $330$ ones for validation. 
	Finally, we follow the similar procedure to project them onto the sphere and create equirectangular images as we do for MNIST-012 dataset.

For our first set of experiments we train the network in~\cite{bb:TIGraNet} with the following parameters. We use two spectral convolutional layers with 10 and 20 filters correspondingly, with global pooling which selects $P_1$ and $P_2$ nodes, where the parameters $P_1=2000$ and $P_2=200$ for MNIST-012 dataset and $P_1=2000$ , $P_2=700$ for ETH-80 dataset. We then use a statistical layer with $12\times 2$ statistics and three fully-connected layers with ReLU and 500, 300, 100 neurons correspondingly.

We have evaluated our approaches with respect to baseline methods in terms of classification accuracy. The \textbf{MNIST-012} dataset is then primarily used for the analysis of both the architecture and the graph construction approach. We then report the final comparisons to state-of-the-art approaches on the \textbf{ETH-80} dataset.

\begin{figure}
	\centering
	\begin{tabular}{cccc}
		\hspace{-0.2cm}&\hspace{-0.6cm}  Input & \hspace{-0.6cm} $\mathcal{F}_0(y)$ & \hspace{-0.6cm} $\mathcal{F}_1(y)$\\
		\vspace{-0.3cm} 	
		\raisebox{0.9cm}0 &
		\hspace{-0.6cm} 	
		\includegraphics[width=0.25\linewidth]{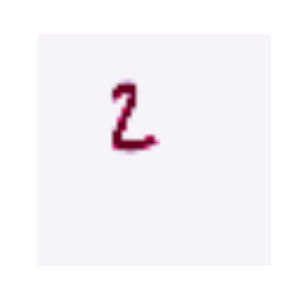} &
		\hspace{-0.6cm} 	
		\includegraphics[width=0.25\linewidth]{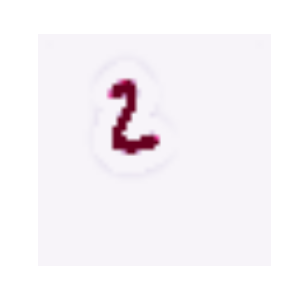} &
		\hspace{-0.6cm} 
		\includegraphics[width=0.25\linewidth]{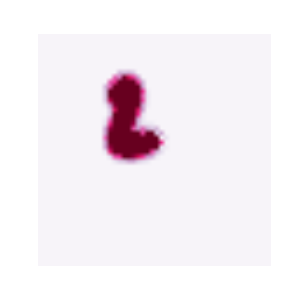} \\
		\vspace{-0.3cm} 	
		\raisebox{0.9cm}{$\frac{1}{8}$} &
		\hspace{-0.6cm} 	
		\includegraphics[width=0.25\linewidth]{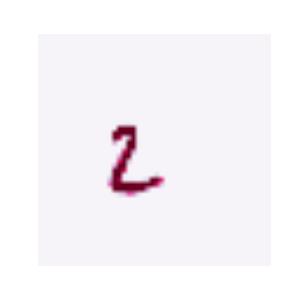} &
		\hspace{-0.6cm} 
		\includegraphics[width=0.25\linewidth]{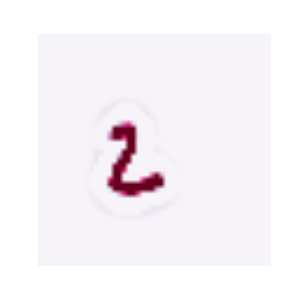} &
		\hspace{-0.6cm} 
		\includegraphics[width=0.25\linewidth]{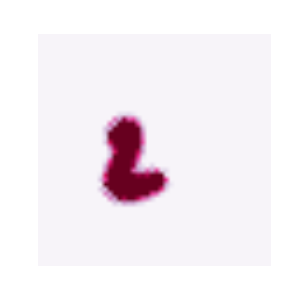} \\
		\vspace{-0.3cm} 	
		\raisebox{0.9cm}{$\frac{1}{4}$}  &
		\hspace{-0.6cm} 	
		\includegraphics[width=0.25\linewidth]{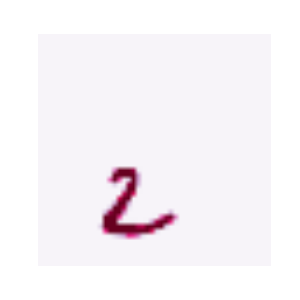} &
		\hspace{-0.6cm} 
		\includegraphics[width=0.25\linewidth]{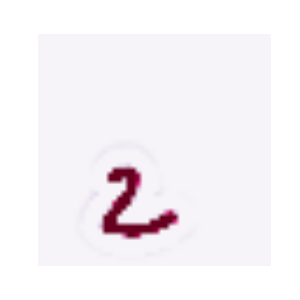} &
		\hspace{-0.6cm} 
		\includegraphics[width=0.25\linewidth]{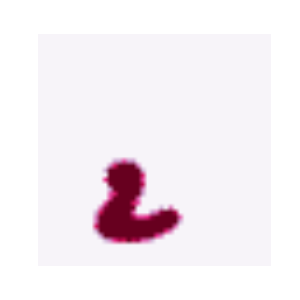} \\
		\hspace{+0.05cm} 
	\end{tabular}
	\caption{Example of the feature maps of the last spectral convolutional layer extracted from equirectangular images. The first column corresponds to the original images created for the same object, which is projected from different tangent planes $\bT_i$, with $\phi_i \in \{0, \frac{1}{8}, \frac{1}{4}\}$; the last two columns show the feature maps given by two randomly selected filters.}
	\label{fig:example_fm}
\end{figure}
\begin{figure}
	\centering
	\begin{tabular}{cccc}
		\hspace{-0.2cm}&\hspace{-0.6cm}  Input & \hspace{-0.6cm} $\mathcal{F}_0(y)$ & \hspace{-0.6cm} $\mathcal{F}_1(y)$\\
		\vspace{-0.3cm} 	
		\raisebox{0.9cm}0 &
		\hspace{-0.6cm} 	
		\includegraphics[width=0.25\linewidth]{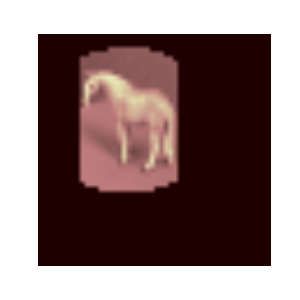} &
		\hspace{-0.6cm} 	
		\includegraphics[width=0.25\linewidth]{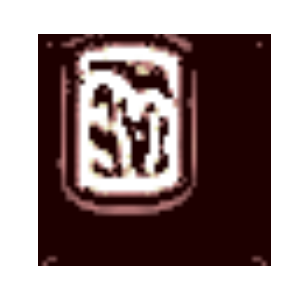} &
		\hspace{-0.6cm} 
		\includegraphics[width=0.25\linewidth]{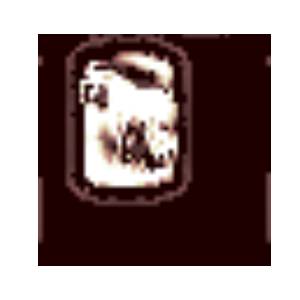} \\
		\vspace{-0.3cm} 	
		\raisebox{0.9cm}{$\frac{1}{8}$} &
		\hspace{-0.6cm} 	
		\includegraphics[width=0.25\linewidth]{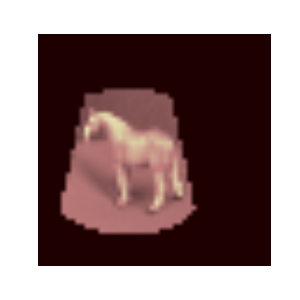} &
		\hspace{-0.6cm} 
		\includegraphics[width=0.25\linewidth]{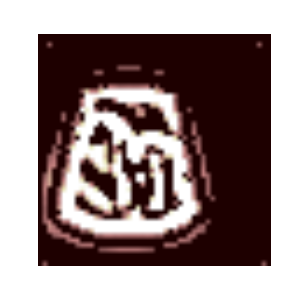} &
		\hspace{-0.6cm} 
		\includegraphics[width=0.25\linewidth]{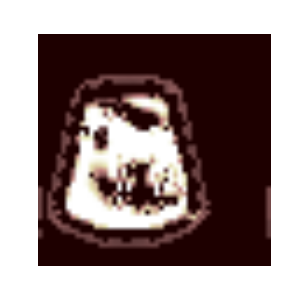} \\
		\vspace{-0.3cm} 	
		\raisebox{0.9cm}{$\frac{1}{4}$}  &
		\hspace{-0.6cm} 	
		\includegraphics[width=0.25\linewidth]{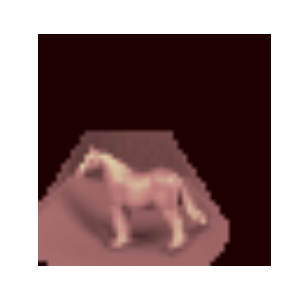} &
		\hspace{-0.6cm} 
		\includegraphics[width=0.25\linewidth]{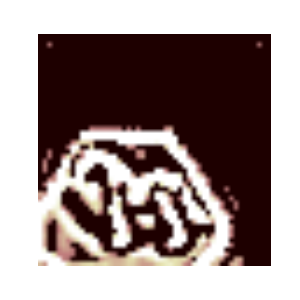} &
		\hspace{-0.6cm} 
		\includegraphics[width=0.25\linewidth]{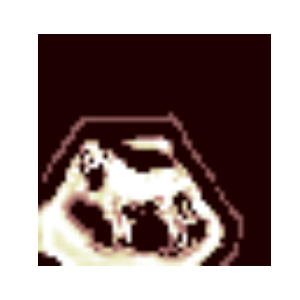} \\
		\hspace{+0.05cm} 
	\end{tabular}
	\caption{Example of feature maps of equirectangular images from \textbf{ETH-80} datasets. Here, we randomly select an input image from the test set and project it on three elevation $\phi_i \in \{0, 1/8, 1/4\}$ and two spectral filters, which are named $\mathcal{F}_1$ and $\mathcal{F}_2$. The figure illustrates resulting feature maps given by the selected filters (second and third columns) and input images (first column).}
	\label{fig:res_eth}
\end{figure}

First of all, we visually show that feature maps on the last convolutional layer of our network are similar for different positions, namely for different tangent planes $\bT_i$ with the same object. 
Fig.~\ref{fig:example_fm} and ~\ref{fig:res_eth}  depict some feature maps of images from MNIST-012 and ETH-80 correspondingly. 

The first column of each figure shows original equirectangular images of the same object projected to different elevations $\phi_i=[0,1/8,1/4]$ and the rest visualize feature maps produced by two randomly selected filters. We can see, that the feature maps stay similar independently of the distortion of the corresponding input image. 
We believe that this, further, leads to closer feature representations, which is essential for good classification.


\begin{figure}
	\centering
	\begin{tabular}{cc}
		\includegraphics[width=0.44\linewidth]{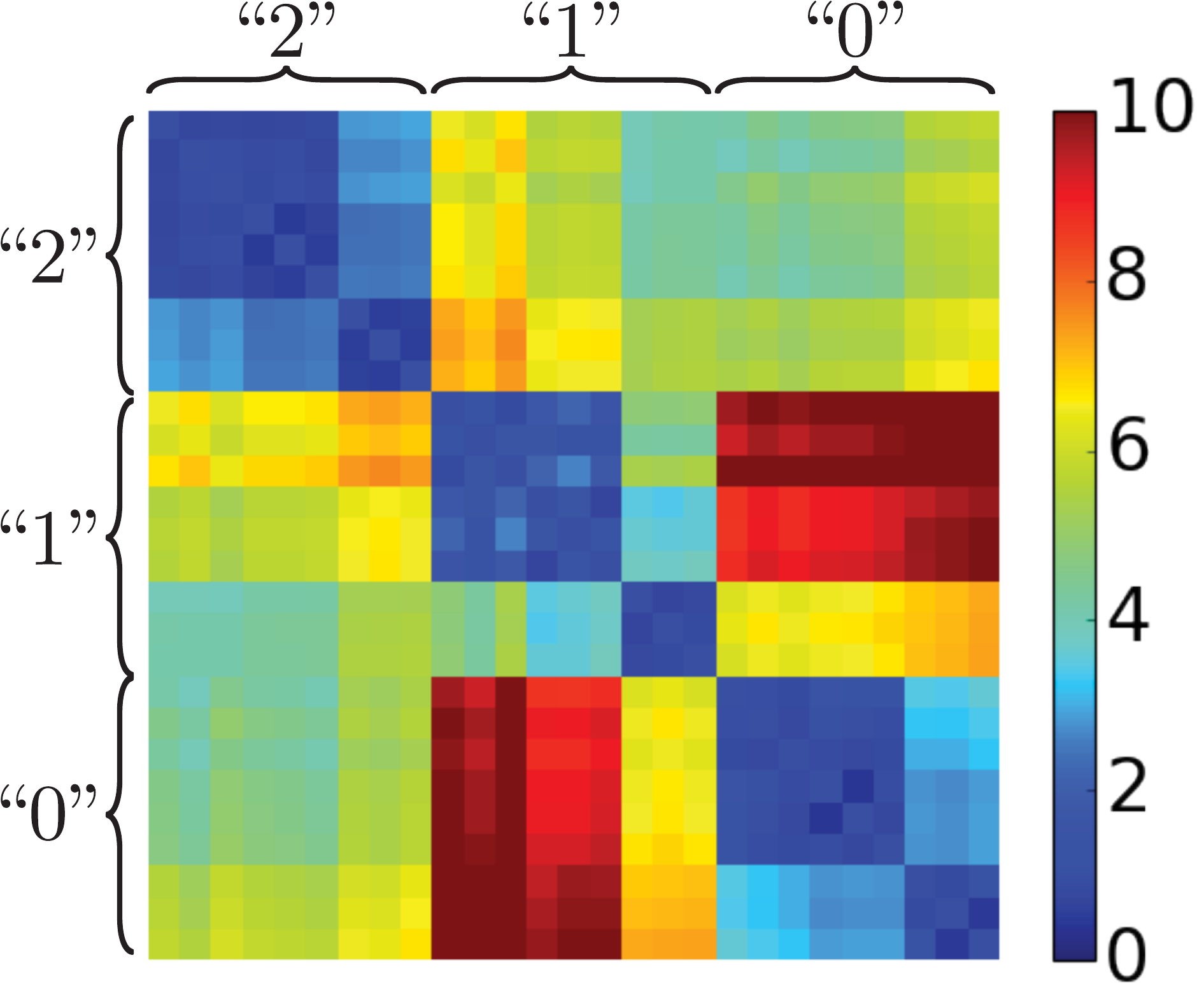} &
		\includegraphics[width=0.44\linewidth]{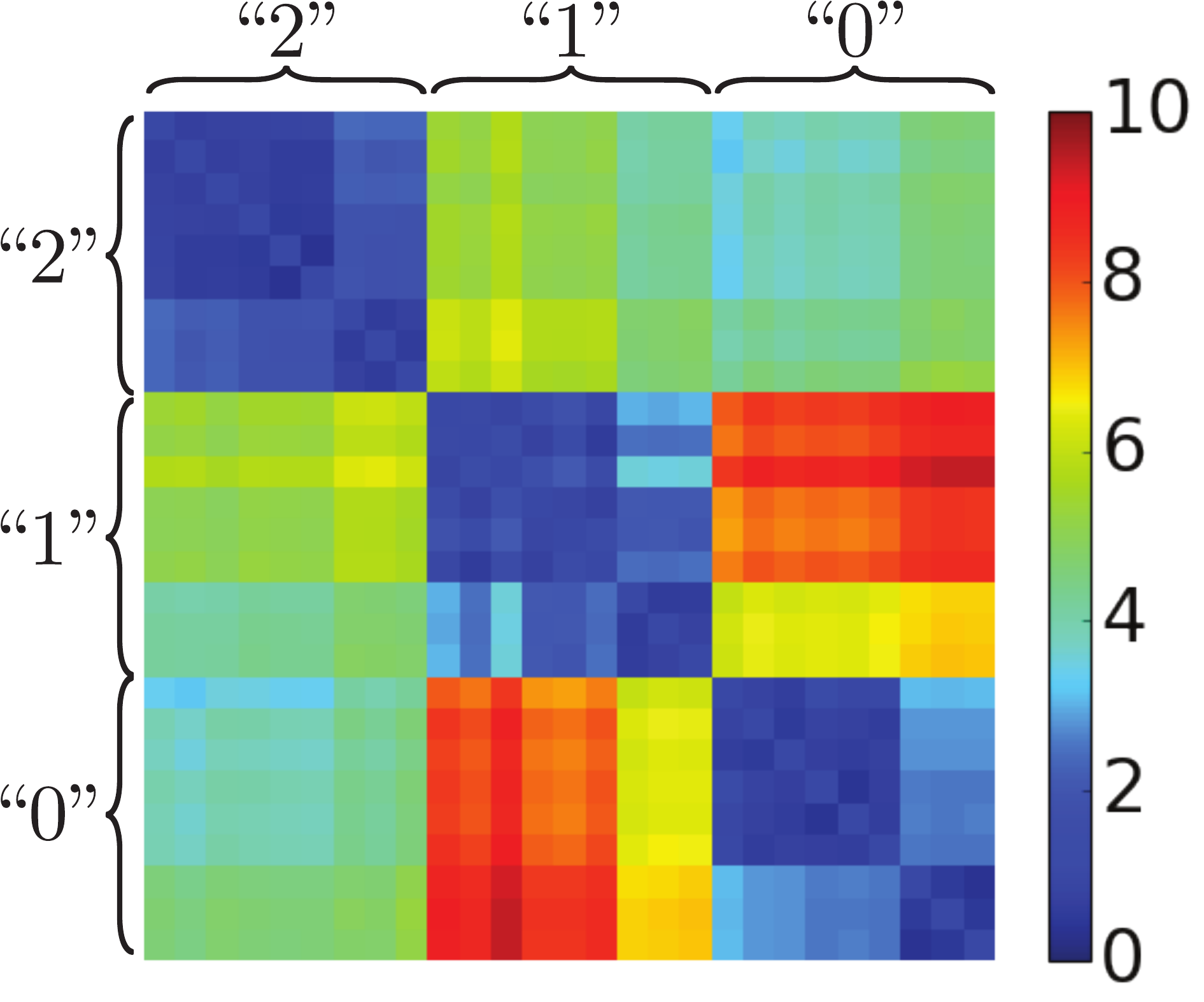} \\
		a) & b) \\
	\end{tabular}
	\caption{Illustration of the Euclidean distances between the features given by the networks of a) \cite{bb:TIGraNet} and b) our geometry-aware graph. 
		This figure depicts resulting matrix $[27\times27]$ for the images from 3 classes and 9 different positions, where axises correspond to the image indexes.  
		Each diagonal $[9\times9]$ sub-matrix corresponds to the same object (digits ``2", ``1", ``0"), the lowest value (blue) corresponds to the most similar features and the highest value (red) to the least similar (best seen in color).}
	\label{fig:res_euc}
\end{figure}

We recall that the goal of our new method is to construct a graph to process images from omnidirectional camera and use it to create similar feature vectors for the same object for different positions~$(\phi_i, \theta_i)$ of the tangent plane. To justify the advantage of proposed approach we design the following experiment. First of all, we randomly select three images of digits `2', `1' an `0' from the test set of MNIST-012. We then project each of these images to 9 positions on the sphere $\phi_i \in \{0, 1/8, 1/4\}, \theta_i \in \{\pm 1/8, 0\}$. 
We then evaluate Euclidean distances between the features that are given by the statistical layer of the network for all pairs of these 27 images. Fig.~\ref{fig:res_euc} presents the resulting $[27 \times 27]$ matrix of this experiment for a grid graph and for the proposed graph representation, which captures the lens geometry. Ideally we expect that images with the same digit give the same feature vector regardless of different elevations $\phi_i$ of the tangent plane. This essentially means that cells of the distance matrix should have low value on the $[9 \times 9]$ diagonal sub-matrices, which correspond to the same object, and high values on the rest of the matrix elements. Fig.~\ref{fig:res_euc} shows that our method gives more similar features for the same object compared to the approach based on a grid graph. This suggests that building graph based on image geometry, as described in Section 4.2, makes features less sensitive to image distortions. This consequently simplifies the learning process of the algorithm.

\begin{table}[t!]
	\centering
	\begin{tabularx}{\linewidth}{X|c|c|c}
		\toprule
		& &  & \textbf{ETH-80} \\
		\scriptsize{method} & \scriptsize{graph type} & \scriptsize{$\#$ Parameters} & \multicolumn{1}{c}{\scriptsize{Accuracy (\%)}} \\
		\midrule
		\scriptsize{\textit{classic Deep Learning:}} &&& \\
		\quad FC Nets & -- & ~1.4M &  71.3 \\
		\quad STN~\cite{bb:STN} & -- &~1.1M & 73.1 \\	
		\quad ConvNets~\cite{bb:ConvNet} & -- &~1.1M & 76.7 \\	
		\scriptsize{\textit{graph-based DLA:}} &&& \\
		\quad ChebNet~\cite{bb:Mikhael}& grid & ~3.8M & 72.9\\
		\quad TIGraNet~\cite{bb:TIGraNet}& grid & ~0.4M & 74.2\\
		\quad ChebNet~\cite{bb:Mikhael}& geometry & ~3.8M & 78.6\\
		\midrule
		Ours & geometry & ~0.4M &  \textbf{80.7} \\ 
		\bottomrule
	\end{tabularx}
			\hspace{+0.05cm} 
	\caption{Comparison to the state-of-the-art methods on the \textbf{ETH-80} datasets. We select the architecture of different methods to feature similar number of convolutional filters and neurons in the fully-connected layers.}
	\label{tab:comparison1}
\end{table}

We further evaluated our approach with respect to the state-of-the-art  methods on ETH-80 dataset. The competing deep learning approaches can be divided in classical and graph-based methods. Among the former ones we use Fully-connected Networks (FCN), Convolutional Network (ConvNets)~\cite{bb:ConvNet} and Spatial Transformer Networks (STN)~\cite{bb:STN}. STN has an additional to ConvNets layer which is able to learn specific transformation of a given input image.  
Among the graph-based methods, we choose ChebNet~\cite{bb:Mikhael} and TIGraNet~\cite{bb:TIGraNet} for our experiments. ChebNet is a network designed based on Chebyshev polynomial filters. TIGraNet is a method invariant to isometric transformation of the input signal.
The architectures are selected such that the number of parameters in convolutional and  fully-connected layers roughly match each other across different techniques. More precisely, all networks have $2$ convolutional layers with $10$ and $20$ filters, correspondingly, and $3$ fully-connected layers with $300, 200$ and $100$ neurons. Filter size of the convolutional layer in classical architectures is $5\times5$. For ChebNet we try polynomials of degree $5$ and $10$ and pick the latter one as it produces better results. For TIGraNet we use polynomial filters of degree $5$.  
The results of this experiment are presented in Table~\ref{tab:comparison1}. 

Table~\ref{tab:comparison1} further shows that ConvNet~\cite{bb:ConvNet} outperforms TIGraNet~\cite{bb:TIGraNet}. This likely happens as  \cite{bb:TIGraNet} gathers global statistics and loses the information about the location of the particular object. This information, however, is crucial for the network to adapt to different distortions on omnidirectional images. We can see that the introduced graph construction method helps to create  similar feature representations for the same object at different elevations, which results into different distortion effects but similar feature response. Therefore, the object looks similar for the network and global statistics become more meaningful compared to a method based on the regular grid graph~\cite{bb:TIGraNet}.

Further, we can see that the proposed graph construction method allows to improve accuracy of both graph-based algorithms: ChebNet-geometry outperforms ChebNet-grid, and proposed algorithm based on TIGraNet outperforms the same method on the grid-graph~\cite{bb:TIGraNet}. Finally, we also notice that our geometry-based method performs better than ChebNet-geometry on the ETH-80 task due to the isometric transformation invariant features; these are an advantage for the image classification problems, where images are captured from different viewpoints. 

Thus, we can conclude that our algorithm produces similar filter responses for the same object at different positions. This, in combination with global graph-based statistics, leads to the better classification accuracy.  

\section{Conclusion}
\label{s:conc}

In this paper  we propose a novel image classification method based on deep neural network that is specifically designed for omnidirectional cameras, which introduce severe geometric distortion effects.
Our graph construction method allows learning filters that respond similarly to the same object seen at different elevations on the equirectangular image. We evaluated our method on challenging datasets and prove its effectiveness in comparison to state-of-the-art approaches that are agnostic to the geometry of the images.

Our discussion in this paper was limited to specific type of the mapping projection. However, the proposed solution has a potential to be extend to more general geometries of the camera lenses. 

	
\clearpage
	

{\small
	\bibliographystyle{ieeetr}
	\bibliography{string,ref}
}

\end{document}